%% file: main.tex
\documentclass[10pt,twocolumn,letterpaper]{article}

 \usepackage{cvpr}              % To produce the CAMERA-READY version
\input{preamble}

\title{\vspace{-1.3cm}\small \textbf{ } \\ \textmd{Accepted at 6th Workshop and Competition on Affective Behavior Analysis in-the-wild \\ CVPR 2024 Workshop} \\ \vspace{5mm} \LARGE \bf 3D Human Pose Estimation with Occlusions: Introducing BlendMimic3D Dataset and GCN Refinement}

\author{Filipa Lino,\ 
 Carlos Santiago,\  
Manuel Marques\\
Institute for Systems and Robotics, LARSyS, Instituto Superior Técnico, Portugal\\
{\tt\small \{filipa.lino, carlos.santiago\}@tecnico.ulisboa.pt, manuel@isr.tecnico.ulisboa.pt }
}
\usepackage{array}
\newcolumntype{?}{!{\vrule width 1pt}}
\usepackage{xcolor}
\definecolor{thecolor}{rgb}{0.1,0.5,1}

\usepackage{fancyhdr}

\begin{document}
\maketitle
\input{sec/0_abstract} 
\input{sec/1_intro}

\input{sec/2_relatedwork}
\input{sec/3_blendmimic3D} 
\input{sec/4_poserefinement}
\input{sec/5_setup}
\input{sec/6_results}
\input{sec/7_conclusion}

%\noindent\textbf{Acknowledgements}.
{\small
\section*{Acknowledgements}
This work was supported by research grant 10.54499/2022.07849.CEECIND/CP1713/CT0001 and LARSyS funding (DOI: 10.54499/LA/P/0083/2020, 10.54499/UIDP/50009/2020, 10.54499/UIDB/50009/2020), through Fundação para a Ciência e a Tecnologia, and by the SmartRetail project [PRR - C645440011-00000062], through IAPMEI - Agência para a Competitividade e Inovação.}

{
    \small
    \bibliographystyle{ieeenat_fullname}
    \bibliography{main}
}


\input{sec/X_suppl}

\end{document}

%% file: preamble.tex
\usepackage[dvipsnames]{xcolor}
\usepackage[accsupp]{axessibility} 
\definecolor{cvprblue}{rgb}{0.21,0.49,0.74}
\usepackage{color, colortbl}
\definecolor{Gray}{gray}{0.9}
\usepackage[pagebackref,breaklinks,colorlinks,citecolor=cvprblue]{hyperref}
\usepackage{amssymb,pifont}

%% file: sec/0_abstract.tex
\begin{abstract}
In the field of 3D Human Pose Estimation (HPE), accurately estimating human pose, especially in scenarios with occlusions, is a significant challenge. 
This work identifies and addresses a gap in the current state of the art in 3D HPE concerning the scarcity of data and strategies for handling occlusions. 
We introduce our novel BlendMimic3D dataset, designed to mimic real-world situations where occlusions occur for seamless integration in 3D HPE algorithms. Additionally, we propose a 3D pose refinement block, employing a Graph Convolutional Network (GCN) to enhance pose representation through a graph model. This GCN block acts as a plug-and-play solution, adaptable to various 3D HPE frameworks without requiring retraining them. By training the GCN with occluded data from BlendMimic3D, we demonstrate significant improvements in resolving occluded poses, with comparable results for non-occluded ones. Project web page is available at \url{https://blendmimic3d.github.io/BlendMimic3D/}. %Upon acceptance, the dataset, code and project page will be released.

%\begin{keywords}
%3D Human Pose Estimation (HPE), Occlusion Handling, BlendMimic3D Dataset, Graph Convolutional Network (GCN), Pose Refinement.
%\end{keywords}
\end{abstract}

%% file: sec/1_intro.tex
\section{Introduction}
\label{sec:intro}

Human pose estimation (HPE) from visual data has become crucial in computer vision, with wide-ranging applications from sports analysis to enhancing smart retail experiences. It involves interpreting a person's position and orientation from images or videos. Despite the emergence of various techniques~\cite{martinez2017simple,tome2017lifting,VideoPose3D,Spatio-Temp} and datasets~\cite{6682899,SURREAL,AMASS}, 3D HPE remains challenging, particularly with monocular camera views in occluded scenarios. Under occlusions, estimating 3D poses becomes even harder, due to the increased ambiguity, and the lack of datasets specifically targeting occluded scenarios makes most state-of-the-art approach struggle with this type of data.

\begin{figure}[t]
      \centering
      \includegraphics[width=1\linewidth]{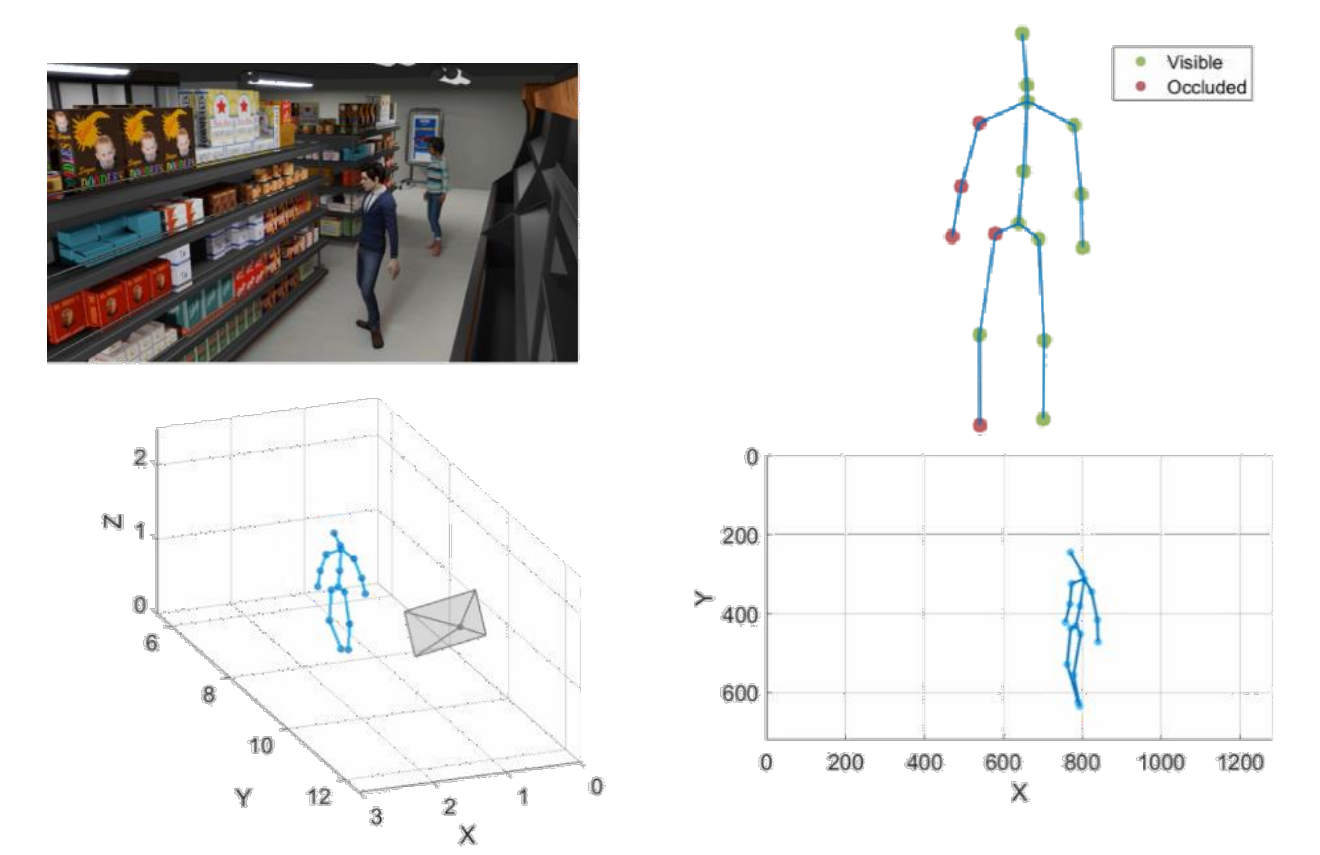}
      \caption{ BlendMimic3D, our synthetic dataset for 3D HPE occlusion benchmarking, features diverse multi-camera scenarios with up to three subjects. It includes Blender animations (top left), keypoint visibility (top right), cameras' parameters, 3D poses (bottom left) and 2D pose representations (bottom right).}
      \label{fig:blenderData}
   \end{figure}

To fill this gap, we introduce a new synthetic dataset, called BlendMimic3D\footnote{Available at https://github.com/FilipaLino/BlendMimic3D}, illustrated in Figure~\ref{fig:blenderData}, that aims to serve as a novel benchmark for HPE with occlusions. Our dataset, built using Blender~\cite{blender_off}, comprises a variety of scenarios mirroring real-world complexities, and purposely contains several types of occlusions, including self, object-based and out-of-frame occlusions. This makes BlendMimic3D an invaluable tool for both training HPE models and benchmarking their performance in occluded scenarios.

Additionally, this work proposes a new pose refinement module\footnote{Available at https://github.com/FilipaLino/GCN-Pose-Refinement}, designed to overcome the limitations of the current state of the art. Our approach is based on a graph convolutional network (GCN)~\cite{GCN} that takes into account spatial and temporal information and is compatible with various 2D-to-3D HPE backbones, including VideoPose3D~\cite{VideoPose3D}, PoseFormerV2~\cite{PFV2}, and D3DP~\cite{D3DP}. It works as a plugin feature that enhances occluded keypoint estimates and does not require training or fine-tuning the HPE backbone, substantially simplifying its use.

The main contributions are the following:
\begin{itemize}

    \item BlendMimic3D dataset: a comprehensive, realistic synthetic benchmark dataset focused on occlusions, aiding in the training and evaluation of HPE models.
    
    \item A novel GCN for 3D pose refinement, leveraging spatial-temporal keypoint relationships. It integrates with most current monocular 3D HPE methods and is designed to address occlusions without additional retraining.

\end{itemize}

Extensive evaluation with two different 2D keypoint detection algorithms~\cite{wu_kirillov_massa_lo_girshick_2019,CPN} and three state-of-the-art 2D-to-3D algorithms~\cite{VideoPose3D,D3DP,PFV2}, validate the utility of our new benchmark dataset and the efficacy of our proposed refinement module in estimating poses in occluded conditions.

%% file: sec/2_relatedwork.tex
\section{Related Work}
\label{sec:related work}
Advances in deep learning, especially Convolutional Neural Networks (CNNs)~\cite{CNN}, have significantly improved HPE, offering enhanced accuracy and speed. Toshev et al.'s DeepPose~\cite{toshev2014deeppose} exemplifies the potential of these methods. In HPE, three primary body modeling approaches are used: kinematic (body keypoints)~\cite{COCO,MPII,PoseTrack, joo2015panoptic,6682899, SURREAL,AMASS,arnab2019exploiting}; planar (body contours)~\cite{OcclAware,SURREAL}; and volumetric (3D meshes)~\cite{6682899, SURREAL, arnab2019exploiting, AMASS, BEDLAM}. Our work focuses on the kinematic model due to its versatility.

\subsection{From 2D to 3D Transition }
When estimating 3D human pose from 2D video inputs, direct 3D estimation~\cite{li20153d,pavlakos2018ordinal} is challenging due to the loss of depth information in 2D representations. Instead, researchers have found it more effective to first extract the 2D pose and then infer the corresponding 3D pose~\cite{martinez2017simple,tome2017lifting}. 

Lee and Chen~\cite{lee1985determination} were early pioneers in 2D joints projection into 3D spaces, but the arrival of deep learning later shifted the focus towards neural network-based methods. Martinez et al.~\cite{martinez2017simple} emphasized the critical role of 2D pose data in predicting 3D keypoints. 

Current methods have adopted two primary paradigms: bottom-up~\cite{fang2017rmpe, cao2017realtime} and top-down~\cite{he2017mask,CPN,wu_kirillov_massa_lo_girshick_2019}. While the former starts with individual body joint estimations, the latter begins by detecting persons. Each approach comes with its set of advantages and challenges, with the trade-off between accuracy and computational speed being paramount.

\subsection{2D HPE}
Single-person estimation primarily employs regression methods, such as DeepPose~\cite{toshev2014deeppose}, and heatmap-based techniques~\cite{cao2017realtime, li2018bottom}. Multi-person scenarios see the use of both bottom-up methods, like OpenPose~\cite{cao2017realtime}, and top-down  strategies like AlphaPose~\cite{fang2017rmpe}. Hybrid methods, highlighted by Miaopeng Li et al.~\cite{li2018bottom}, merge these techniques. 

Another noteworthy top-down model in this category is Mask R-CNN by Kaiming He et al.~\cite{he2017mask}, initially designed for object detection and semantic segmentation, but later incorporated HPE. Based on that framework, Detectron2~\cite{wu_kirillov_massa_lo_girshick_2019} was introduced to handle tasks from object detection to 2D keypoint identification. It uses CNNs to generate heatmaps for keypoints, with the heatmap's peak indicating the exact keypoint location for accurate results. Also following~\cite{he2017mask}, Chen et al.~\cite{CPN} developed the Cascaded Pyramid Network (CPN) to improve multi-person pose estimation, focusing on ``hard" keypoints that are occluded or not visible.

%\subsection{Tracking}
%When considering multi-person scenarios, consistent pose estimation of a specific individual requires effective human tracking. The Simple Online and Realtime Tracking (SORT)~\cite{SORT} method by Bewley et al. lays the groundwork with a two-stage process: initial feature extraction followed by object classification in proposed regions.

%Built on SORT, DeepSort~\cite{DeepSort} incorporated a CNN model to focus on human-related image features. This integration allows the model to adeptly handle challenging scenarios such as occlusions and appearance variations. Moreover, this model specializes in person re-identification, ensuring consistent tracking of individuals over time. 

Our study employed Detectron2~\cite{wu_kirillov_massa_lo_girshick_2019} and CPN~\cite{CPN} for 2D keypoint detection due to their precision and state-of-the-art features, with both achieving a high performance on the COCO~\cite{COCO} benchmark. We also integrated DeepSort~\cite{DeepSort} for tracking individuals in multi-person scenarios, basing 3D pose predictions from specific 2D keypoints.

\subsection{3D HPE }
Despite advances in 2D HPE, 3D HPE struggles with depth ambiguities, limited datasets, and complexities associated with occlusions. Considering monocular RGB images and videos and a two-stage approach (2D to 3D Lifting), Martinez et al.~\cite{martinez2017simple} set a benchmark in this domain by using a fully connected residual network to regress 3D joint locations from 2D ones. Another influential work by Tome et al.~\cite{tome2017lifting} proposed a multi-stage approach where 2D and 3D poses are processed concurrently.

Additionally, temporal data from videos has been incorporated to address depth issues. Pavllo et al.~\cite{VideoPose3D} introduced a temporal dilated convolutional model, named VideoPose3D. While this approach is noted for its simplicity and efficiency, it may encounter difficulties in handling continuous occlusions.

Motivated by that, Cheng et al.~\cite{OcclAware} presented a network that addressing occlusions through temporal frame analysis. This architecture is particularly effective in scenarios with occluded body parts, but only accounts for self-occlusions, since the testing was conducted using data that primarily featured such occlusions, limiting its applicability.

Zheng et al.~\cite{PF} introduced PoseFormer, a purely transformer-based model for 3D HPE from videos. This model process both spatial and temporal aspects of human movement. Building on this, Zhao et al.~\cite{PFV2} developed PoseFormerV2, which employs the frequency domain to boost efficiency and accuracy of 3D HPE. This approach reduces computational demands and increases robustness to noisy in 2D joint detections, making it effective in complex and occluded scenarios. 

 Shan et al.~\cite{D3DP} presented D3DP, an innovative method for probabilistic 3D HPE. D3DP generates multiple potential 3D poses from a single 2D observation, using a denoiser conditioned on 2D keypoints to refine the poses. The hypotheses for the 3D poses are reprojected onto the 2D camera plane, and the best hypothesis for each joint is selected based on reprojection errors. These selections are combined to form the final pose.

\subsection{Graph Convolutional Network}
Graph-based approaches, such as GCNs~\cite{GCN}, can be used to address occlusions in 3D HPE by representing the body as a graph, where each node represents a body keypoint and each edge represents the relationship between two joints. A notable application is the Dynamic Graph Convolutional Network (DGCN) introduced by Zhongwei Qiu et al.~\cite{DGCN}, that can model relationships between 2D joints over time. Wenbo Hu et al.~\cite{directGCN} proposed representing a 3D human skeleton as a directed graph, to capture hierarchical orders among the joints. 

Following the DGCN approach, to further enrich the 3D HPE domain, Cai et al.~\cite{Spatio-Temp} proposed a graph-based approach leveraging spatial-temporal relationships. They formulated 2D pose sequences as graphs and designed a network to capture multi-scale features and temporal constraints. Later, Yu Cheng et al.~\cite{Graph&TempCNN} presented a novel framework for estimating 3D multi-person poses from monocular videos with two directed GCNs, one dedicated to joints and the other to bones, which together estimate the full pose. This framework integrates GCNs and Temporal Convolutional Networks (TCNs)~\cite{TCN} to handle challenges like occlusions and inaccuracies in person detection. They also include directed graph-based joint and bone GCNs.

Our proposal employs a GCN model, which is designed to represent the 3D human pose as an enhanced, undirected graph, inspired by the method in~\cite{Spatio-Temp}. We have tailored our model to specifically refine 3D pose predictions, particularly effective in scenarios with occlusions. This is achieved by expanding joint relationships, with training conducted on a variety of cases involving occlusions.  

\subsection{HPE Datasets}

The evolution of HPE approaches has underscored the importance of comprehensive datasets, especially in the context of occlusions. For 2D HPE, datasets like MPII~\cite{MPII}, COCO~\cite{COCO}, and PoseTrack~\cite{PoseTrack} offer diverse scenarios ranging from static images to dynamic videos, capturing real-world complexities. These datasets facilitate the development of models that generalize to multiple environments.

Well-known 3D HPE datasets, such as Human3.6M~\cite{6682899}, SURREAL~\cite{SURREAL}, and AMASS~\cite{AMASS}, typically require sophisticated equipment like motion capture systems for accurate pose recording. While these datasets offer high precision, they often face challenges in diversity and real-world applicability. For multi-person scenarios, datasets like CMU Panoptic~\cite{joo2015panoptic}, 3DPW\cite{arnab2019exploiting} and AGORA~\cite{agora} become crucial as they capture more complex interactions and dynamics, including occlusions. Table~\ref{tab:Datasets} illustrates the diversity and focus of some of these datasets.

\begin{table}[t]
    \centering
    % \caption{Summary of 3D HPE Datasets. Data type `R' and `S' denote `Real' and `Synthetic'. \(^\dagger\): focus on non-self occlusions, including object-based, multi-person, and out-of-frame types. \(^\ddagger\):~indicate keypoint visibility and additional metadata.}
    \caption{3D HPE Datasets. Data type `R' and `S' denote `Real' and `Synthetic'. \(^\dagger\) non-self occlusions (object-based, multi-person, and out-of-frame). \(^\ddagger\) annotations of keypoint visibility.}
    \scriptsize % Reducing the font size
    \setlength\tabcolsep{1.5pt} % Adjust horizontal padding
  \begin{tabular}{l |c c c c c c c|}
        
        \cline{2-8}
         \multicolumn{1}{c|}{}&\textbf{Data} & \multicolumn{1}{c}{\textbf{No. of}}&\textbf{ Action} & \textbf{Single-} & \textbf{Multi-} & \multicolumn{2}{c|}{\textbf{Occlusions}}\\
        % \cline{1-3}
        \textbf{Dataset} & \textbf{Type} & \textbf{ Frames} & \textbf{Tags} &  \textbf{Person}&\textbf{Person} & \textbf{Complex}\(^\dagger\) & \textbf{Labels}\(^\ddagger\) \\
        \specialrule{.13em}{.05em}{.05em}
       \rowcolor{Gray} Human3.6M~\cite{6682899} & R & 3.6M & \checkmark & \checkmark & \ding{53} & \ding{53} & \ding{53} \\
       CMU Panoptic~\cite{joo2015panoptic} & R & 1.5M & \ding{53} & \checkmark & \checkmark & \ding{53} & \ding{53}  \\
        \rowcolor{Gray} SURREAL~\cite{SURREAL} & S & 6M & \ding{53} & \checkmark & \ding{53} & \ding{53} & \ding{53}  \\
        3DPW~\cite{arnab2019exploiting}& R & 51K & \ding{53}&\checkmark &\checkmark &\checkmark &\ding{53}   \\
        \rowcolor{Gray} AGORA~\cite{agora} & S & 17K & \ding{53} & \ding{53} & \checkmark & \checkmark & \ding{53} \\
        BEDLAM~\cite{BEDLAM} & S & 380K & \ding{53} &\checkmark &\checkmark &\checkmark & \ding{53}\\
         \hline
        \rowcolor{Gray} \textbf{BlendMimic3D } & \textbf{S} &\textbf{136K} & \checkmark & \checkmark& \checkmark& \checkmark & \checkmark  \\
    \specialrule{.13em}{.05em}{.05em}
    \end{tabular}
    \label{tab:Datasets}
\end{table}

Following the discussion on existing datasets, the introduction of the BEDLAM dataset~\cite{BEDLAM} represents a significant advancement. As a synthetic dataset designed for 3D human pose and shape (HPS) estimation, BEDLAM demonstrated that neural networks trained solely on synthetic data can achieve state-of-the-art accuracy in 3D HPS estimation from real images.

While both the COCO~\cite{COCO} and Human3.6M~\cite{6682899} datasets have been instrumental in advancing state-of-the-art algorithms, they present limitations. COCO's human-curated nature is prone to errors, whereas Human3.6M, although providing high-precision pose data, lacks in representing occluded scenarios. Wandt et al.~\cite{H36MOccl} showed that state-of-the-art 3D HPE models significantly underperform when faced with synthetic occlusions.

Addressing the challenges in 3D HPE occlusion handling highlighted by Wandt et al., we introduce BlendMimic3D. Inspired by Human3.6M and leveraging BEDLAM's synthetic capabilities, BlendMimic3D offers advanced occlusion management across various levels. It sets a new benchmark in occlusion-aware 3D HPE with action-oriented labeled activities and occlusions, marking keypoints' visibility per frame as shown in Table~\ref{tab:Datasets}.

%In response to the need for better occlusion handling in 3D HPE, and addressing the challenges identified by Wandt et al.~\cite{H36MOccl}, we introduce BlendMimic3D. Drawing inspiration from Human3.6M's structure and BEDLAM's synthetic achievements, BlendMimic3D is engineered to encompass a broad spectrum of occlusion levels, setting a new standard in occlusion-aware 3D HPE datasets. Table~\ref{tab:Datasets} shows that, like Human3.6M, BlendMimic3D is action-oriented with labeled activities, uniquely featuring labeled occlusions to indicate the visibility of each keypoint in every frame.

\addtolength{\textheight}{-3cm} 

%% file: sec/3_blendmimic3D.tex
\begin{figure*}[t]
      \centering
      \includegraphics[width=\linewidth]{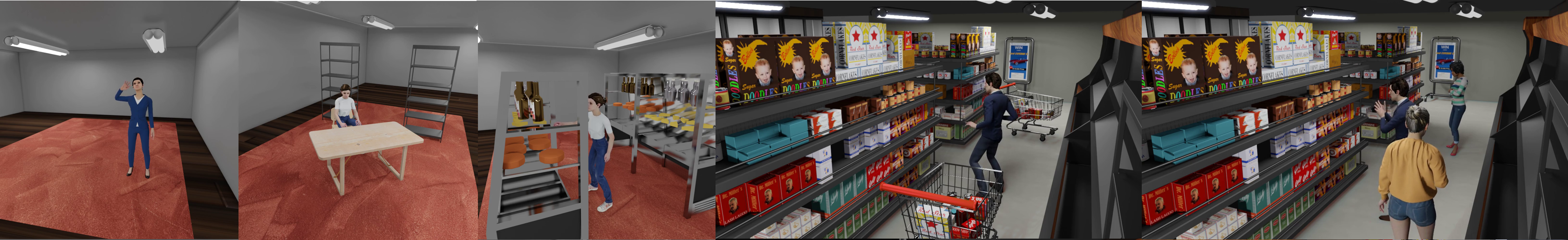}
      \caption{Visual representation of different scenes from BlendMimic3D datasets. From left to right: synthetic subjects, SS1, SS2 and SS3.}
      \label{fig:h36m&blendmimc}
\end{figure*}

\section{BlendMimic3D Dataset}
As the need for HPE grows, so does the demand for detailed datasets to train and test models. The efficacy of these datasets is judged by their accuracy, completeness, and variety. Creating 3D HPE datasets is complex and usually requires special tools such as MoCap systems and wearable devices, resulting in datasets created in controlled settings. As argued by Wandt et al.~\cite{H36MOccl}, despite the progress achieved with Human3.6M~\cite{6682899} dataset, there remains a notable gap that synthetic datasets can address.
\begin{figure}[t]
      \centering
      \includegraphics[width=\linewidth]{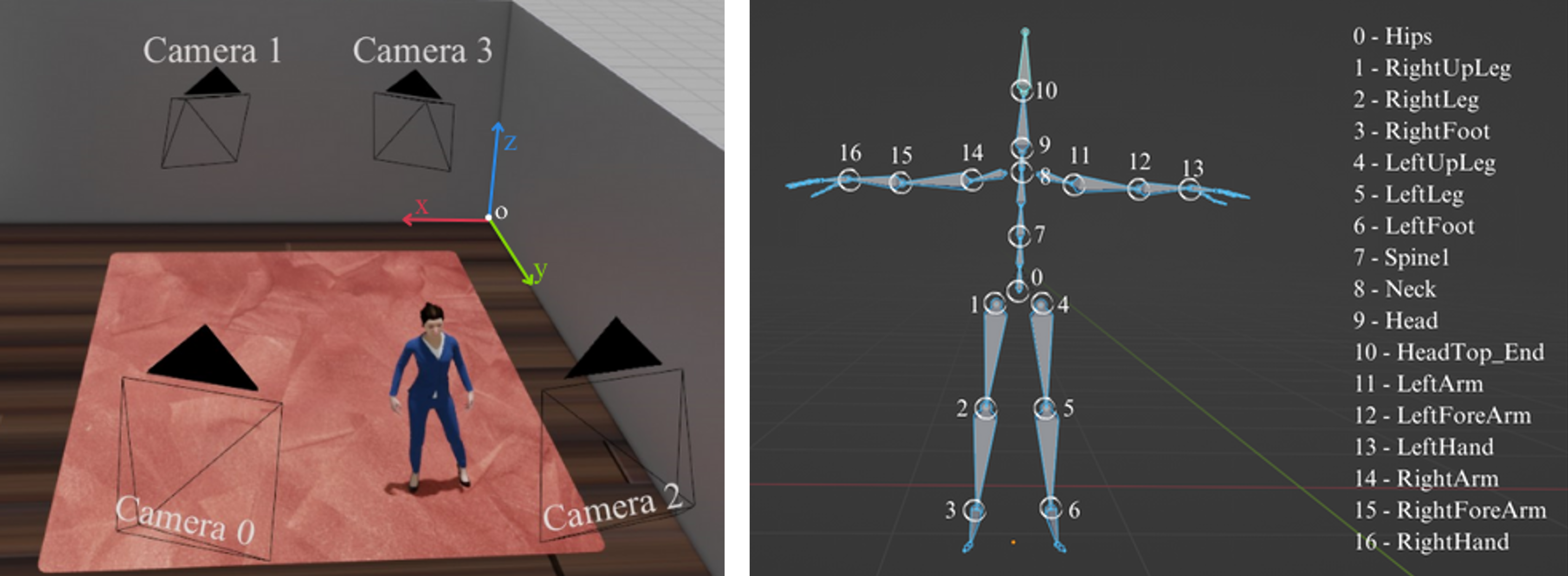}
      \caption{Left: Camera distribution with the world coordinate system at the origin, with subject SS1 of BlendMimic3D dataset. Right: Visualization of 3D character armature, highlighting the specific keypoints used for coordinate extraction.}
      \label{fig:blender}
\end{figure}

Using Blender~\cite{blender_off}, a popular open-source 3D computer graphics software, we introduce a synthetic dataset tailored to address challenges such as self, object-based and out-of-frame occlusions. To ensure its adaptability and relevance, we designed it to comprise a diverse set of scenarios, from simple environments resembling Human3.6M~\cite{6682899}, to more complex ones with numerous occlusions and multi-person contexts. Figure~\ref{fig:h36m&blendmimc} illustrates examples of frames from videos in our dataset, showcasing the range of settings and multi-person contexts of BlendMimic3D. More detailed examples are available in the supplementary material.

In the process of crafting BlendMimic3D, four cameras were positioned within the virtual environment, as depicted in Figure~\ref{fig:blender}~(Left). A skeletal framework, shown in Figure~\ref{fig:blender}~(Right), was attached to a 3D character model, enabling animation of our synthetic subjects. Utilizing resources from Mixamo\footnote{\url{https://www.mixamo.com/\#/}}~\cite{mixamo}, the characters were animated to simulate a range of actions, such as ``Arguing", ``Greeting", or ``Picking Objects". From each camera's perspective, videos were generated utilizing Blender's rendering engine. The resulting dataset comprises:
\begin{enumerate}
    \item 3 scenarios, from a simple environment to more complex and realistic ones;
    \item 3 subjects, each one performing several different actions; %(SS1 and SS2 performed 14 different actions each and SS3 performed 2);
    \item Single and multi-person settings, with up to 3 subjects;
    \item A total of 128 videos with an average duration of 35 seconds (1050 frames).
\end{enumerate} 
Metadata is available for all videos, including the parameters used for camera calibration, 2D and 3D positions of keypoints, as well as a binary array depicting which keypoints were occluded in each frame. All this extracted data is illustrated in Figure~\ref{fig:blenderData}.

BlendMimic3D is organized in the same manner as the Human3.6M dataset, with videos categorized by subject and action. The dataset includes synthetic subjects designated as SS1, SS2 and SS3. While SS1 focuses on self-occlusions, SS2 addresses object and out-of-frame occlusions. Each of these subjects covers 14 distinct actions. SS3, set in a smart store environment, manages both occlusions and multi-person scenarios. It offers two variations of the same action—one in a single-person context and the other in a multi-person setting. Just like with Human3.6M, each synthetic action is captured in four videos, each with a different perspective. 

%% file: sec/4_poserefinement.tex
\section{Pose Refinement with GCN}

\begin{figure}[t]
      \centering
      \includegraphics[width=\linewidth]{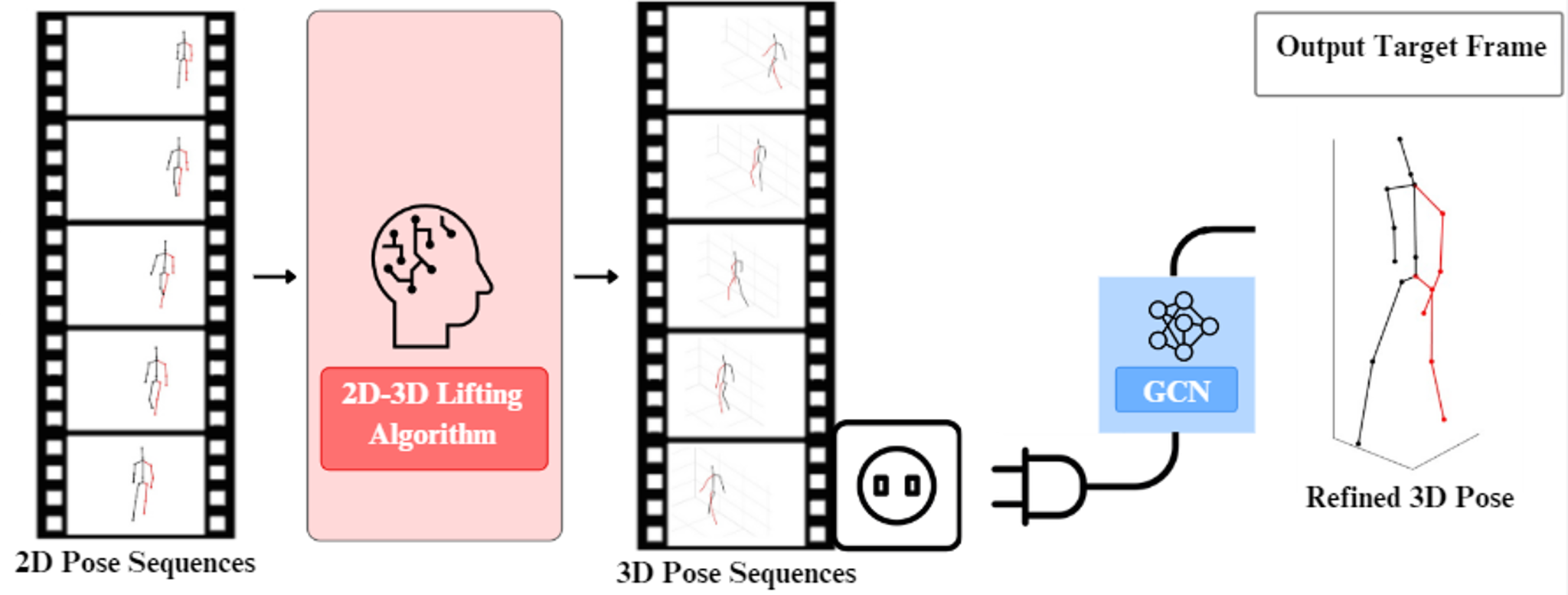}
      \caption{Overview of the proposed framework. After any chosen 3D HPE algorithm, our Graph Convolutional Network (GCN) refines the estimated 3D poses by integrating spatial and temporal insights, leading to enhanced and precise 3D pose estimation, particularly effective in handling occlusions.}
      \label{fig:GCNPlugin_pt2}
    \end{figure}

Our proposed methodology is a pose refinement stage, illustrated in Figure~\ref{fig:GCNPlugin_pt2}, where we introduce our Graph Convolutional Network (GCN) as a plugin to enhance the estimated 3D poses. Our GCN is trained on BlendMimic3D dataset, which provides a diverse range of occlusion scenarios. This allows the network to learn and adapt to various occlusion types, refining the pose estimation for occluded joints.

\begin{figure}[t]
      \centering
      \includegraphics[width=0.5\linewidth]{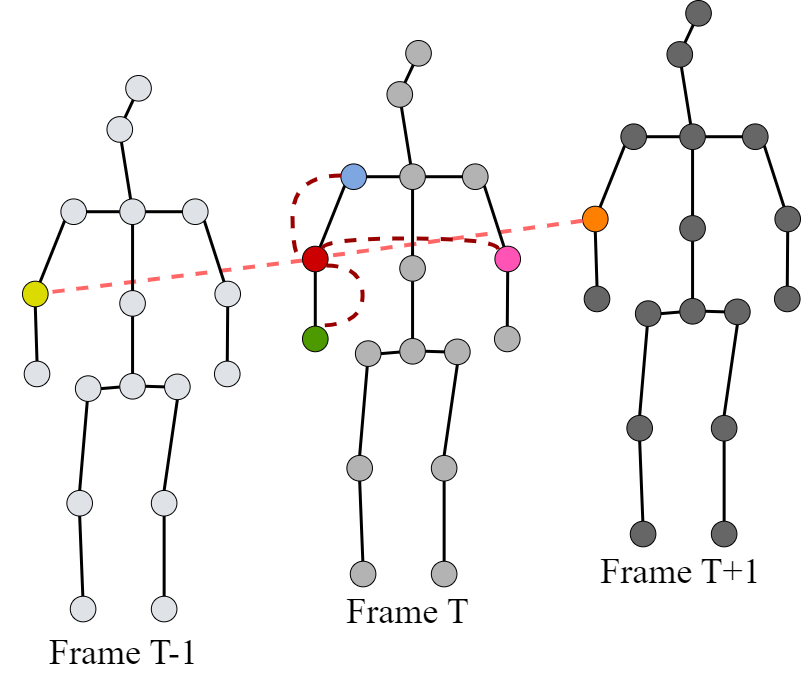}
      % \caption{Illustration of the graph dynamics. This example highlights the considered neighboring nodes for the right elbow keypoint, categorized into six classes: (1) Center (red). (2) Physically-connected node closer to the spine (blue). (3) Physically-connected farther from the spine (green). (4) Symmetric node (pink). (5) Time-forward node (orange). (6) Time-backward (yellow). }
      \caption{Illustration of the graph dynamics for the right elbow keypoint, with neighboring nodes categorized into six classes: (1) Center (red). (2) Physically-connected node closer to the spine (blue). (3) Physically-connected farther from the spine (green). (4) Symmetric node (pink). (5) Time-forward node (orange). (6) Time-backward (yellow). }
      \label{fig:GCN}
   \end{figure}

The GCN not only considers spatial relationships between body joints but also temporal continuity across frames. It conceptualizes the human body as a graph structure, where nodes are body keypoints and edges represent joint connections. Unlike traditional models that primarily link a keypoint to its immediate neighbors~\cite{directGCN}, our model, inspired by the work of Cai et al.~\cite{Spatio-Temp} and Yu Cheng et al.~\cite{Graph&TempCNN}, establishes broader connections across consecutive frames, as depicted in Figure~\ref{fig:GCN}. This extended connectivity is crucial for accurately inferring occluded or ambiguous keypoints in challenging environments.

Drawing on the formulation by Kipf and Welling~\cite{GCN}, their GCN approach refines spectral graph convolutions to enhance efficiency and scalability. Given a graph \( \mathcal{G} \) with an adjacency matrix \( A \), the propagation rule in their GCN model for each layer is expressed as

\begin{equation}
H^{(l+1)} = \sigma \left( \tilde{D}^{-\frac{1}{2}} \tilde{A} \tilde{D}^{-\frac{1}{2}} H^{(l)} W^{(l)} \right),
\label{eq:kipf}
\end{equation}
where \( H^{(l)} \) is the activation matrix for the \( l \)-th layer. \( H^{(0)} \) denotes the input feature matrix, with each row representing a feature vector for every node. \( W^{(l)} \), often referred to as the kernel, is the weight matrix for the \( l \)-th layer. \( \sigma \) is an activation function, typically the ReLU. The augmented adjacency matrix, \( \tilde{A} = A + I \), includes self-connections, and  \( \tilde{D} \) is its corresponding diagonal node degree matrix.

Kipf and Welling's strategy uses the normalized adjacency matrix to spread node features across the graph. Normalization by the degree matrix  \( \tilde{D} \) ensures stable gradients and effective training. In (\ref{eq:kipf}), the kernel \( W^{(l)} \), is shared by all 1-hop neighboring nodes, suggesting a consistent treatment of these immediate neighbors.  

To enhance this approach, we expanded from merely considering 1-hop neighbors, recognizing the need for distinct kernels tailored to different neighboring nodes based on their semantics. Following (\ref{eq:kipf}), we devised a spatial-temporal undirected graph \(\mathcal{G} = (\mathcal{V}, \mathcal{E}, A)\). In this graph, \(\mathcal{V}\in \mathbb{R}^{T\times J}\) signifies the vertices set corresponding to \(T\) consecutive frames (one for each, past, present, and future), with \(J\) joints in each frame. \(\mathcal{E}\) represents the nodes' connections. The adjacency matrix \(A \in \mathbb{R}^{P\times P}\), considering \(P=TJ\), that \(a_{ij}=0\) if \((i,j)\not\in \mathcal{E}\) and \(a_{ij}=1\) if \((i,j)\in \mathcal{E}\). This adjacency matrix captures both spatial and temporal dynamics across frames.

By classifying neighboring nodes and understanding their semantic relationships (as illustrated in Figure~\ref{fig:GCN}), we apply distinct kernels for each class of neighborhood. Drawing from~\cite{Spatio-Temp}, consider an input signal \(X \in \mathbb{R}^{P\times C}\) that represents \(C\)-dimensional features of \(P\) vertices on the graph. The convolved signal matrix \(Z \in \mathbb{R}^{P\times C}\), is given by the graph convolution, articulated as

\begin{equation}
    Z = \sum_{k} D_k^{-\frac{1}{2}} A_k D_{k}^{-\frac{1}{2}} X W_k,
    \label{eq:GCN}
\end{equation}
in which \(k\) indexes the neighbor class, \(W_k\) denotes the filter matrix for the \(k\)-th type of 1-hop neighboring nodes. In relation to the normalized \(\tilde{A}= A + I_P\) from equation~(\ref{eq:kipf}), expression~(\ref{eq:GCN}) decomposes it into \(k\) sub-matrices with \(\tilde{A}=\sum_kA_k\). Here, \(D_k^{ii} = \sum_j A_k^{ij}\) represents the degree matrix that normalizes \(A_k\).

\begin{figure}[t]
      \centering
      \includegraphics[width=\linewidth]{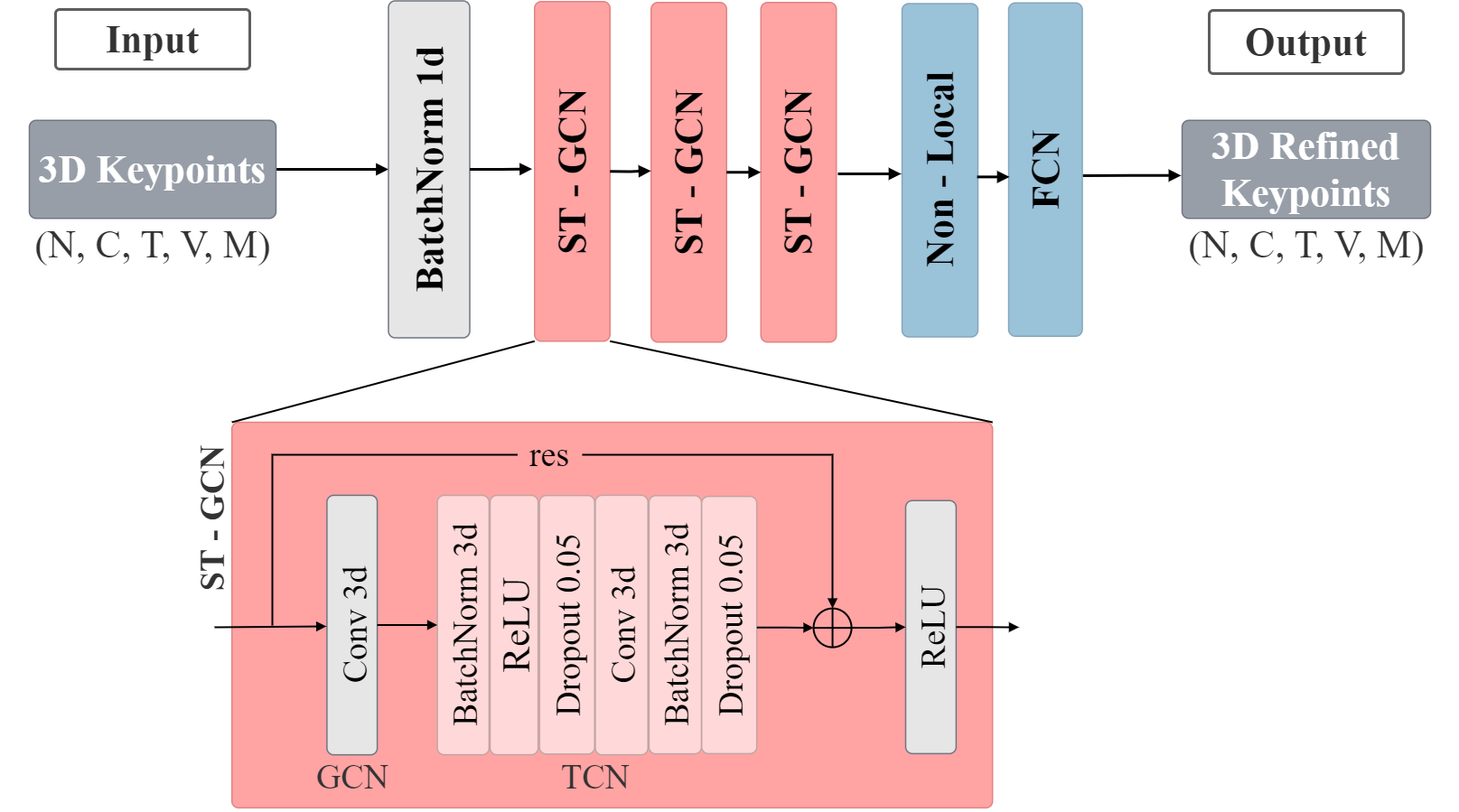}
      \caption{Graph-based 3D human pose refinement architecture with detailed architecture of the Spatial-Temporal Graph Convolutional (ST-GCN) layer. }
      \label{fig:MainGCN}
   \end{figure}

Our model combines graph convolution operation from ~(\ref{eq:GCN}) and 3D convolutions. The primary architecture, depicted in Figure~\ref{fig:MainGCN}, merges both spatial and temporal graph convolutions. The input, a tensor representing 3D keypoints, undergoes normalization for stability. This input tensor has dimensions (N,C,T,V,M), with N as the batch size, C as the number of features, T as the temporal dimension (input sequence length), V as the graph nodes for each frame, and M as the number of instances in a frame.
   
The core of the model comprises several spatial-temporal graph convolutional (ST-GCN) layers, depicted in Figure~\ref{fig:MainGCN} , designed for feature extraction and refinement. A non-local block is also incorporated, capturing long-range dependencies and relationships between different input parts. The resulting features are then passed through a fully connected layer producing the final refined 3D pose.

The details of the ST-GCN layers are shown in Figure~\ref{fig:MainGCN}. Each layer begins with an operation that applies a graph convolution, incorporating the spatial structure and connections defined by an adjacency matrix. Following the graph convolution, the output undergoes a temporal 3D convolution, capturing the temporal relationships across frames. A residual connection is employed to facilitate faster convergence and mitigate the vanishing gradient problem. The final output of each ST-GCN layer passes through a ReLU function for a non-linear transformation. The combination of these operations ensures that our model understands the spatial-temporal dynamics of human actions, enabling accurate 3D pose estimation even in challenging scenarios.

%% file: sec/5_setup.tex
\section{Experimental Setup}

Our architecture provides an end-to-end solution for 3D pose estimation from video, handling occlusions—a common real-world challenge. Our tests on standard benchmarks show it outperforms existing top methods, especially in occluded scenarios, while also maintaining strong performance in standard situations. The experimental framework was implemented using Pytorch~\cite{pytorch}, an Intel(R) Core(TM) i7-8700K CPU @ 3.70GHz
and two NVIDIA GeForce GTX 1080 Ti. 

%Our architecture offers an end-to-end pipeline from detection to 3D pose refinement, effectively deriving 3D poses from videos. Tested on standard benchmarks, it excels in occluded situations over current top methods. Our approach not only maintains performance in standard situations but improves the accuracy in cases of occlusions, a common challenge in real-world applications. 

\subsection{Datasets and Evaluation Metrics}
\label{sec:datasets}
\noindent\textbf{Datasets.} A primary dataset in our study is Human3.6M~\cite{6682899}, which serves as the training foundation for several 3D HPE algorithms~\cite{VideoPose3D,D3DP,PF,PFV2,Graph&TempCNN,OcclAware,Spatio-Temp}. Human3.6M furnishes 3.6 million human poses and corresponding images. Captured in controlled indoor environments, it documents 15 unique actions, with two versions each, from four viewpoints. It is important to note that, due to privacy concerns, data from only 7 subjects is available: S1, S5, S6, S7, S8, S9, S11, totaling 840 videos. For performance evaluation, we sourced all available subjects from Human3.6M~\cite{6682899}. These were combined with our synthetic subjects from BlendMimic3D: SS1, SS2 and SS3. 

In line with established practices in 3D HPE research, as seen in previous works ~\cite{VideoPose3D, D3DP, PF, PFV2, Graph&TempCNN,OcclAware,Spatio-Temp}, we have selected a specific set of subjects for our training and testing phases. For the training of our GCN, we utilize the 3D pose predictions from 6 subjects of Human3.6M, S1, S5, S6, S7, S8 and S11, along with our synthetic subjects SS1, and SS2. We then evaluate the performance of our model on two different subjects, S9 and SS3. Both 3D pose predictions and 3D refined poses are represented in the camera's coordinate system and a single model is used to train all camera views for all actions. 

\noindent\textbf{Evaluation metrics.} Our 3D human pose estimation evaluation harnesses the Mean Per-Joint Positional Error (MPJPE)~\cite{survey}, a metric that calculates the average $\ell_2$-norm difference between estimated and true 3D poses, represented by the equation
\begin{equation}
    \text{MPJPE}=\frac{1}{N}\sum_{i=1}^N \left\| J_i - J_i^* \right\|_2,
\label{eq:MPJPE}
\end{equation}
where \( N \) represents the joint count, and \( J_i \) and \( J_i^* \) denote the true and estimated positions of the \( i_{th} \) joint, respectively.% The Procrustes Aligned MPJPE (P-MPJPE)~\cite{survey} first aligns the predicted pose to the target pose using rigid transformations, such as rotation, scale, and translation. The equation for this metric is 
%\begin{equation}
 %   \text{P-MPJPE} = \frac{1}{N}\sum_{i=1}^N \left\| a J_i R + t - J_i^* \right\|_2,
  %  \label{eq:P-MPJPE}
%\end{equation}
%where \( a \) is the scaling factor, \( R \) is the rotation matrix, and \( t \) is the translation vector. 

\subsection{Implementation Details}
\noindent\textbf{2D HPE.} This stage focuses on accurately capturing the skeletal structure in two dimensions, which lays the foundation for the subsequent 3D pose estimation. To identify subjects in video frames and extract their 2D keypoints, we use two detection algorithms in our 2D HPE process: CPN~\cite{CPN} and Detectron2~\cite{wu_kirillov_massa_lo_girshick_2019}. For keypoints detection using Detectron2~\cite{wu_kirillov_massa_lo_girshick_2019}, we utilize a pretrained model that uses Mask R-CNN~\cite{he2017mask} with ResNet-101-FPN~\cite{FPN} as backbone. Regarding CPN~\cite{CPN}, which is an extension of FPN as suggested by~\cite{VideoPose3D}, we employ the 2D keypoint predictions provided from their fine-tuned CPN model for the Human3.6M dataset. For our synthetic 2D pose predictions, we re-implement CPN, using a ResNet-101 backbone with a 384$\times$288 resolution. This model uses externally provided bounding boxes generated by Detectron2.

\begin{figure}[t]
      \centering
      \includegraphics[width=\linewidth]{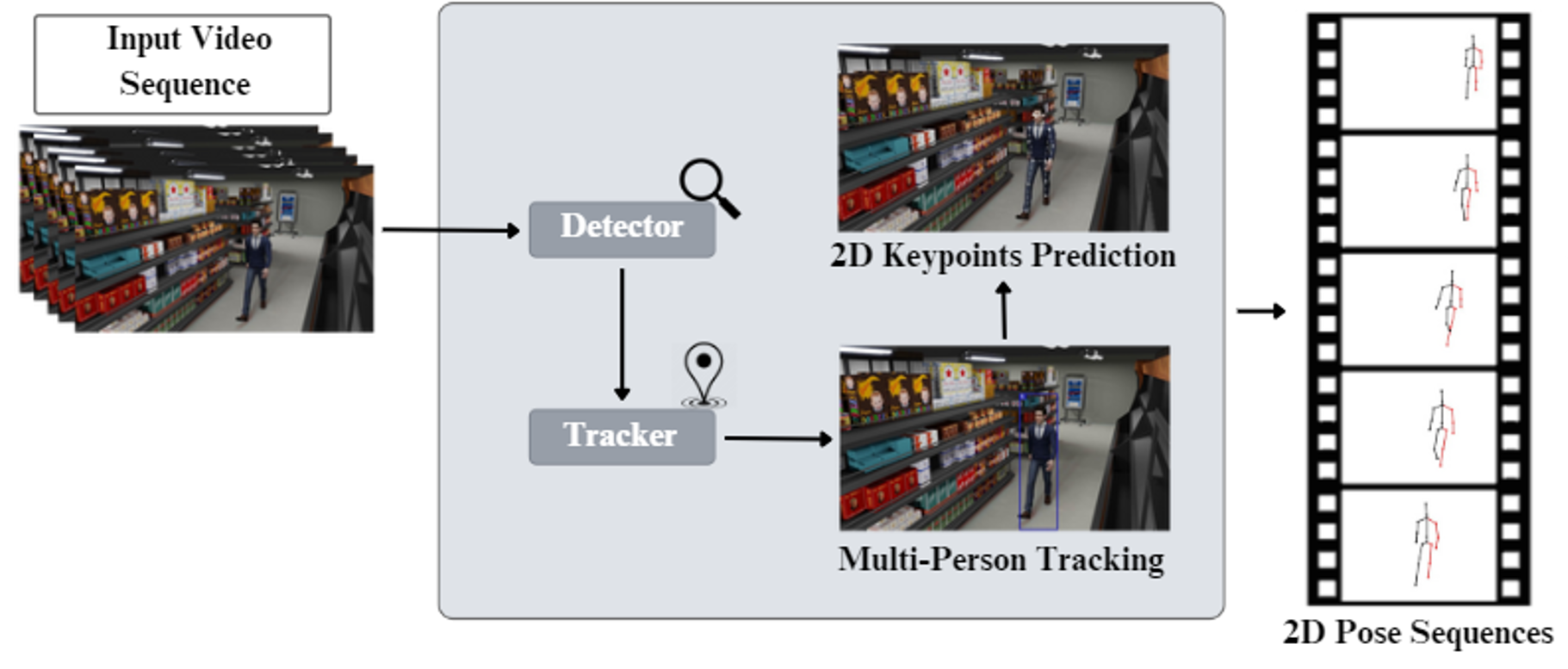}
      \caption{Overview of the proposed preprocessing strategy for 2D HPE. It begins with (1) employing a detection algorithm for pinpointing subjects and capturing their 2D keypoints, followed by (2) a tracking mechanism to maintain focus on a target subject, supplying a sequence of 2D poses. }
      \label{fig:preprocess}
   \end{figure}

To handle dynamic scenes with multiple people, we integrate the DeepSort~\cite{DeepSort} algorithm, modified to track a specific individual with a unique ID. Our method assumes that the individual remains in the frame throughout the monitoring in a multi-person environment. In order to maintain the tracking continuity, especially when the target ID is temporarily lost, we select the closest bounding box based on centroid distance, prioritizing those with high confidence scores. Also, to improve detection performance, our approach resizes subsequent frames according to the previous tracked bounding box and implementing an region of interest (ROI) cropping strategy focused on the target ID. This entire preprocessing approach, encompassing both detection and tracking phases, is illustrated in Figure~\ref{fig:preprocess}.
   
\noindent\textbf{2D-to-3D Pose Conversion} 
We extracted 2D keypoints as inputs for our 2D-to-3D pose lifting module and assessed the performance of various algorithms in handling occlusions with our BlendMimic3D dataset. These algorithms include VideoPose3D~\cite{VideoPose3D}, PoseFormerV2~\cite{PFV2}, and D3DP~\cite{D3DP}, for which we utilized their available pretrained models. For all three algorithms, we used an input sequence length of 243 frames. Specifically, for PoseFormerV2, we inputted 27 frames into the spatial encoder along with 27 DCT coefficients. For D3DP, we configured the model to use 1 hypothesis and one iteration.

%Each algorithm has its strengths in processing temporal and spatial information, crucial for reconstructing the 3D structure of the human body.

\noindent\textbf{GCN Pose Refinement} Our GCN model is trained for 40 epochs with a mini-batch size of 256, using the AMSGrad~\cite{amsgrad} optimizer and an initial learning rate of 0.001. The learning rate is reduced by 0.1 every 5 epochs and shrinks by a factor of 0.95 after each epoch, with a more significant reduction of 0.5 every 5 epochs. Training batches are created by a generator based on pre-split subject IDs for training and testing groups, as detailed in Section~\ref{sec:datasets}. The model is updated through backpropagation based on these batches. We follow the training losses of ~\cite{Spatio-Temp}, including 3D pose loss (MPJPE), derivative loss (measuring the Euclidean distance of the first derivative between predicted and ground truth velocities), and symmetry loss (focusing on the accuracy of left and right bone pairs). Test data is solely used for evaluation. 

%% file: sec/6_results.tex
\section{Experimental Results}

\subsection{Quantitative results}

\begin{figure}[t]
      \centering
      \includegraphics[width=\linewidth]{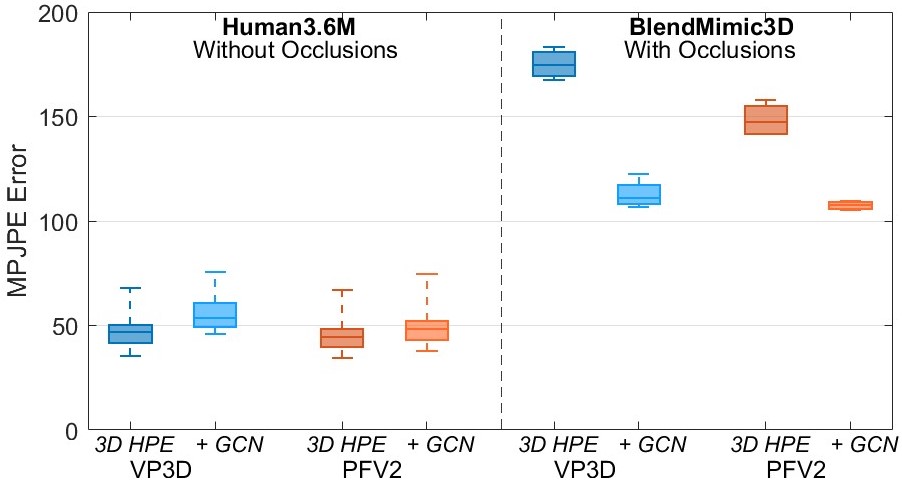}
      \caption{Evaluation of our GCN pose refinement block against previous methods: VideoPose3D~(VP3D) and PoseFormerV2~(PFV2), showcasing performance on CPN-based detections across Human3.6M and BlendMimic3D test sets.}
      \label{fig:Results}
   \end{figure}

Our analysis on Human3.6M and BlendMimic3D datasets, using CPN-based and Detectron2-based 2D detections, demonstrates the positive impact of the GCN pose refinement block in handling occlusions, as evidenced by MPJPE improvements particularly on the occlusion-heavy BlendMimic3D dataset. Figure~\ref{fig:Results} visually summarizes the enhancements and trade-offs introduced by our GCN across VideoPose3D and PoseFormerV2.

Figure~\ref{fig:Results} shows that the results with our GCN achieve a comparable error on the non-occluded Human3.6M dataset. However, the baseline approaches exhibit worse MPJPE in occluded scenarios (BlendMimic3D dataset), while our GCN reduces this error escalation. The proposed approach leads to a notable decrease of more than 30\% on the average errors with occlusions.

Unlike PoseFormerV2 (PFV2) and VideoPose3D (VP3D), D3DP incorporates mechanisms that can handle occlusions, utilizing a diffusion process to add noise and a denoiser conditioned on 2D keypoints, leading to a variety of hypotheses that can capture the possible variations in pose. GCN integration addresses the occlusion management challenges in both VP3D and PFV2 models, as demonstrated in Table~\ref{tab:results}. This table also highlights the GCN's broader impact, including its application to D3DP, within the BlendMimic3D test set.

\begin{table}[t]
    \centering
    \scriptsize % Reducing the font size
    %\caption{Evaluation of our GCN pose refinement block against previous methods on both CPN-based and Detectron2-based detections, utilizing the BlendMimic3D test set. For each 3D HPE algorithm and corresponding 2D detector, the highest scores achieved, both independently and in combination with the GCN, are highlighted in \textcolor{thecolor!40!green}{green}.}
    \caption{Evaluation of our GCN pose refinement block against previous methods, with CPN and Detectron2, on BlendMimic3D test set. Best results are highlighted in \textcolor{thecolor!40!green}{green}.}
    \setlength\tabcolsep{2.5pt} % Adjust horizontal padding

    \begin{tabular}{l?l r|l r|l r}
    %\cline{2-7}
        & \multicolumn{6}{c}{\textbf{3D HPE Model }--\textbf{MPJPE (Avg [mm])} }\\
        %\cline{2-7}
         \textbf{2D HPE}& VP3D~\cite{VideoPose3D} & + GCN & PFV2~\cite{PFV2} & + GCN & D3DP~\cite{D3DP} & + GCN \\
         %\hline
         \specialrule{.13em}{.05em}{.05em}
       \rowcolor{Gray}CPN~\cite{CPN}  & 175.0 & \textcolor{thecolor!40!green}{112.7}&148.6 &\textcolor{thecolor!40!green}{107.5} & 100.7&\textcolor{thecolor!40!green}{95.3}\\
       Detectron2~\cite{wu_kirillov_massa_lo_girshick_2019}  &198.0 &\textcolor{thecolor!40!green}{127.7} &155.0 &\textcolor{thecolor!40!green}{106.9} &99.9 &\textcolor{thecolor!40!green}{95.3}\\
       %\hline
       \specialrule{.13em}{.05em}{.05em}
    \end{tabular}
    \label{tab:results}
\end{table}

Table~\ref{tab:results} underscores the GCN's versatility, showing consistent performance enhancements across different 2D detection methods (CPN and Detectron2). It also shows improvements in D3DP's performance, affirming the GCN's value even in models already equipped for occlusion management. Detailed results, categorized by action for each model, can be found in the supplementary material.

This evaluation highlights BlendMimic3D's role in overcoming the self-occlusion bias of datasets like Human3.6M, emphasizing its importance for enhancing 3D HPE robustness through diverse occlusions. It showcases synthetic data's role in model development and affirms our GCN's advancement in occlusion management, setting a new benchmark for occlusion handling in 3D HPE systems.

\subsection{Qualitative results}
\begin{figure}[t]
      \centering
      \includegraphics[width=0.8\linewidth]{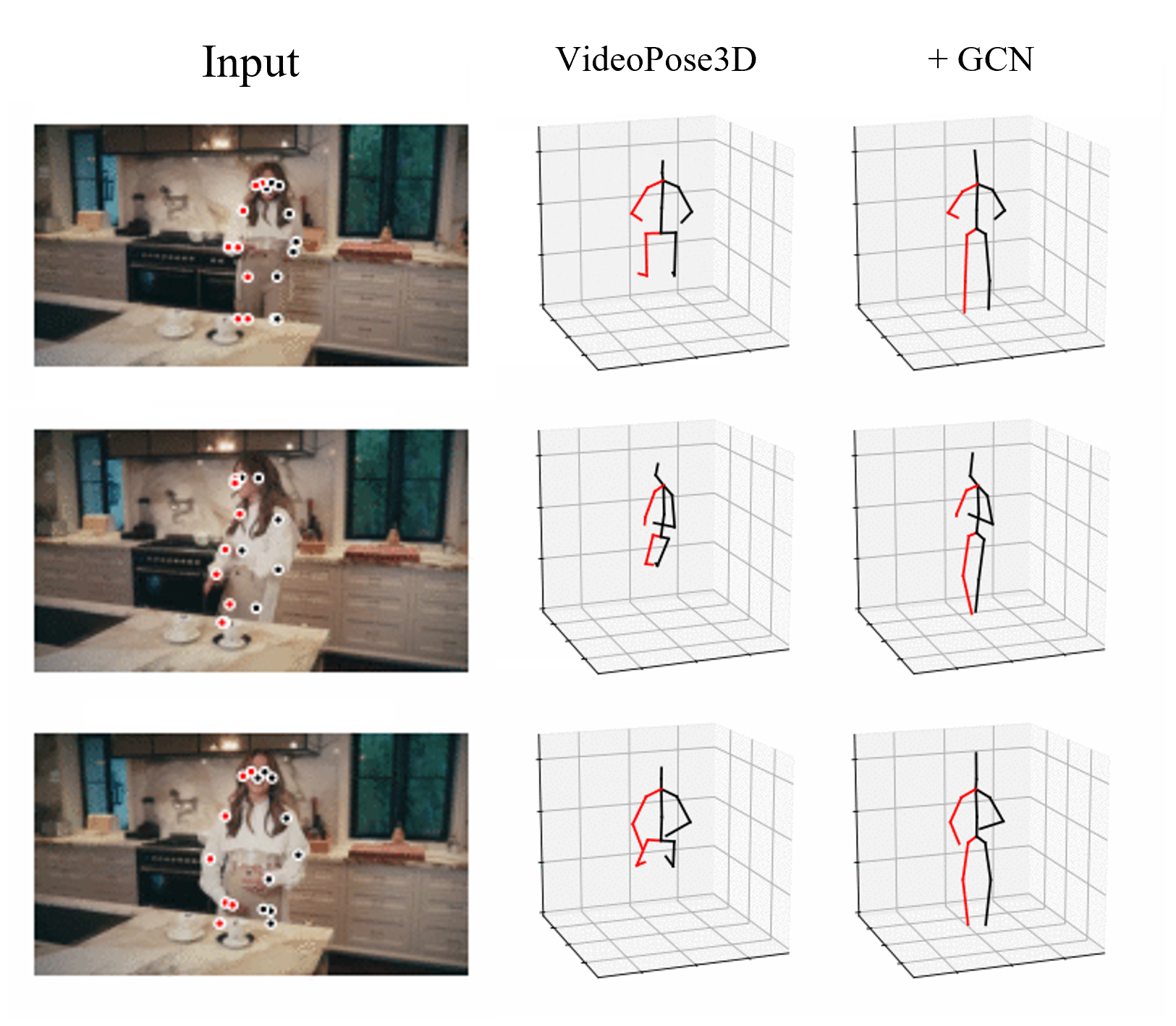}
      \caption{Example showcasing three frames from a real ``in the wild'' video with the corresponding 3D HPE using VideoPose3D, on Detectron2 detections, with and without the proposed GCN.}
      \label{fig:stacked_frames}
   \end{figure}

To test our approach in a real-world scenario featuring occlusions, Figure~\ref{fig:stacked_frames} showcases qualitative results. It compares the 3D human pose estimation from VideoPose3D with and without our refined pose, both derived from the same input video.
As shown in Figure~\ref{fig:stacked_frames}, the GCN approach improves the estimation results, particularly for occluded legs. This suggests that our GCN is effective in handling occlusions -- a critical benefit for real-world applications where such occlusions are frequent.

\addtolength{\textheight}{-3cm} 

%% file: sec/7_conclusion.tex
\section{Conclusion}
This work introduces BlendMimic3D, a new benchmark designed to train and evaluate 3D HPE with occlusions. Unlike traditional datasets such as COCO and Human3.6M, with controlled settings and limited occlusion variations, our BlendMimic3D replicates real-world complexities. A standout feature of BlendMimic3D is its expandability and ease of modification\footnote{https://github.com/FilipaLino/BlendMimic3D-DataExtractor}
, requiring only Blender for animation generation. Additionally, we propose a GCN pose refinement block, that can be plugged in with state-of-the-art 3D HPE algorithms to improve their performance for occluded poses, requiring no further training of the HPE backbone. This ensures that performance improvements in occluded conditions do not compromise accuracy in standard, non-occluded settings. Future efforts will aim to fully preserve performance in these scenarios upon integrating the GCN.

%% file: sec/X_suppl.tex
\clearpage
\setcounter{page}{1}
\maketitlesupplementary

%\cref{sec:intro};

\section{ BlendMimic3D examples }
\begin{figure}[t]
      \centering
      \includegraphics[width=\linewidth]{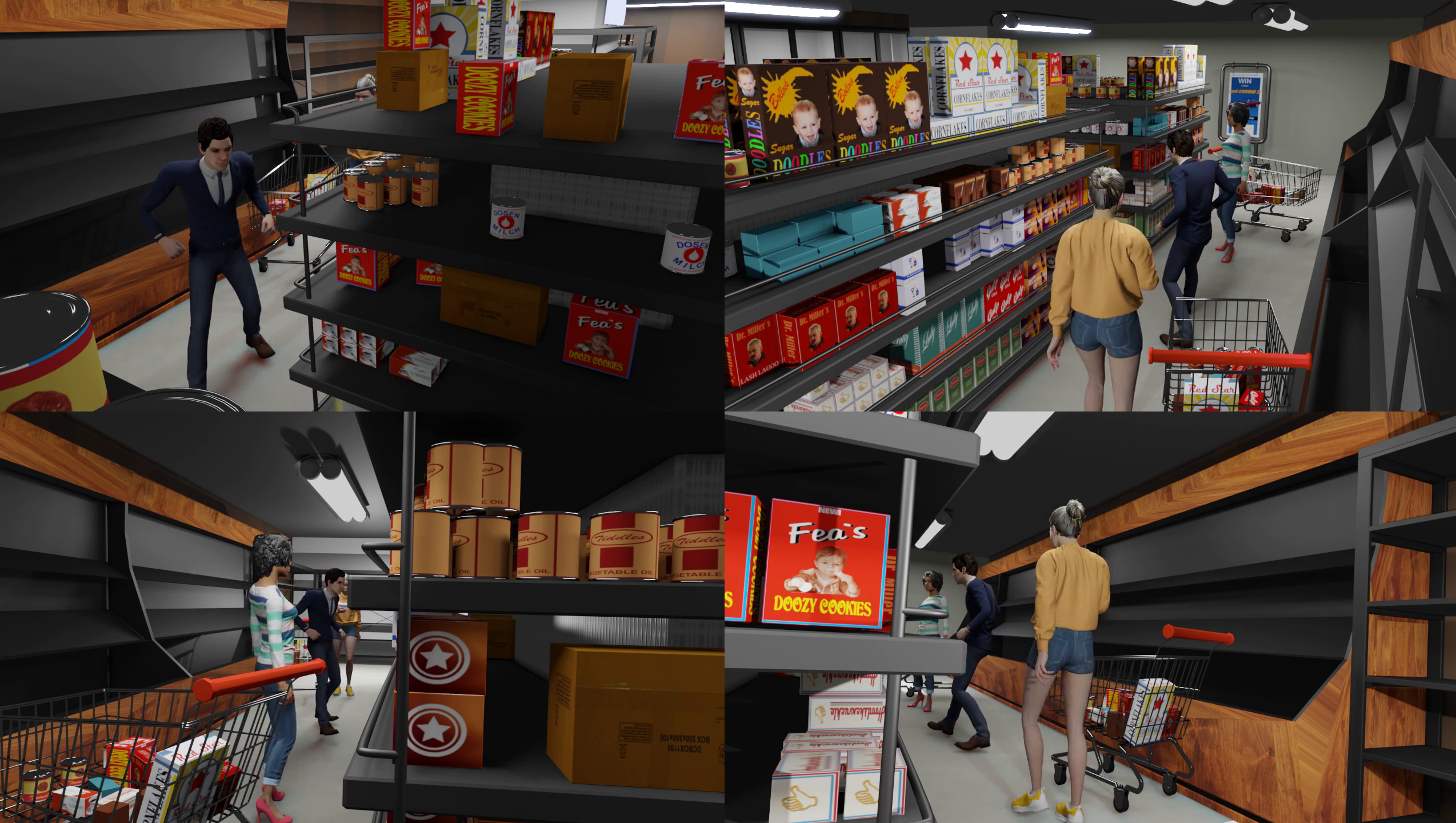}
      \caption{Subject SS3 engaging in the ``Focus\_multi" action. This figure showcases the same frame from different perspectives: (top left) Camera 0; (top right) Camera 1; (bottom left) Camera 2; (bottom right) Camera 3.}
      \label{fig:SS3_examples}
\end{figure}

For illustration purposes, Figure~\ref{fig:SS3_examples} presents all four camera views of the same frame from the ``Focus\_multi" action in our synthetic test set, featuring subject SS3. This particular action is designed to simulate a multi-person scenario within a supermarket setting, where three subjects interact amidst objects, creating occlusions.

\section{Detailed Evaluation of GCN}
Table~\ref{tab:DetailedResults} presents an evaluation of the GCN pose refinement block's performance across actions in the BlendMimic3D test set. Covering both CPN-based and Detectron2-based detections, the table demonstrates MPJPE improvements through the incorporation of the GCN into established 3D HPE models. Highlighted results demonstrate improvements in the accuracy of pose estimation in scenarios with occlusions.

% Table: Model Results
\begin{table*}[t]
\centering
\small
\setlength\tabcolsep{5pt} % Adjust horizontal padding
\caption{Evaluation of our GCN pose refinement block against previous methods. This evaluation includes performance on both CPN-based and Detectron2-based detections, utilizing the BlendMimic3D test set. For each 3D HPE algorithm and corresponding 2D detector, the highest scores achieved, both independently and in combination with the GCN, are highlighted in \textcolor{thecolor!40!green}{green}.}
\label{tab:DetailedResults}
    \begin{tabular}{l l?c c c c |c}
   
    \textbf{Model} & \textbf{2D HPE} & \textbf{TakesItem} [mm] & \textbf{TakesItem\_multi} [mm] & \textbf{Focus} [mm] & \textbf{Focus\_multi} [mm] & \textbf{Avg} [mm] \\
      \specialrule{.13em}{.05em}{.05em}
     \rowcolor{Gray}VideoPose3D~\cite{VideoPose3D} & CPN~\cite{CPN} & 167.5 & 170.8 & 178.3 & 183.4 & 175.0 \\
    + GCN & CPN~\cite{CPN} & \textcolor{thecolor!40!green}{106.9} &\textcolor{thecolor!40!green}{ 109.7} & \textcolor{thecolor!40!green}{112.0} & \textcolor{thecolor!40!green}{122.4} &\textcolor{thecolor!40!green}{ 112.7} \\
    \rowcolor{Gray} VideoPose3D~\cite{VideoPose3D}  & Detectron2~\cite{wu_kirillov_massa_lo_girshick_2019} & 188.8 & 194.6  & 201.8 & 206.7 & 198.0 \\
     + GCN & Detectron2~\cite{wu_kirillov_massa_lo_girshick_2019} & \textcolor{thecolor!40!green}{117.9} & \textcolor{thecolor!40!green}{119.7} & \textcolor{thecolor!40!green}{130.7} & \textcolor{thecolor!40!green}{142.6} & \textcolor{thecolor!40!green}{127.7} \\ \hline
    \rowcolor{Gray} PoseFormerV2~\cite{PFV2} & CPN~\cite{CPN}& 152.6 &157.8 &141.5 &142.2& 148.6\\
      + GCN & CPN~\cite{CPN} & \textcolor{thecolor!40!green}{106.4} &\textcolor{thecolor!40!green}{108.7}& \textcolor{thecolor!40!green}{105.3} &\textcolor{thecolor!40!green}{109.7} &\textcolor{thecolor!40!green}{107.5}\\
    \rowcolor{Gray} PoseFormerV2~\cite{PFV2}  & Detectron2~\cite{wu_kirillov_massa_lo_girshick_2019} & 157.8 & 164.9 & 142.5 & 154.8 & 155.0    \\
     + GCN & Detectron2~\cite{wu_kirillov_massa_lo_girshick_2019} &\textcolor{thecolor!40!green}{ 103.7}& \textcolor{thecolor!40!green}{106.2}& \textcolor{thecolor!40!green}{99.3} & \textcolor{thecolor!40!green}{118.3} & \textcolor{thecolor!40!green}{106.9}\\
    \hline
   \rowcolor{Gray} D3DP~\cite{D3DP} & CPN~\cite{CPN} & 94.7 & 95.9 & 100.3 & 112.1 & 100.7\\ 
    + GCN & CPN~\cite{CPN} & \textcolor{thecolor!40!green}{91.8} & \textcolor{thecolor!40!green}{93.3} & \textcolor{thecolor!40!green}{96.3} &\textcolor{thecolor!40!green}{ 99.9} & \textcolor{thecolor!40!green}{95.3}\\
     \rowcolor{Gray} D3DP\cite{D3DP}&Detectron2~\cite{wu_kirillov_massa_lo_girshick_2019} & \textcolor{thecolor!40!green}{88.4}& 95.0 & 101.7 & 114.6 & 99.9 \\
    + GCN &Detectron2~\cite{wu_kirillov_massa_lo_girshick_2019} & 88.8& \textcolor{thecolor!40!green}{94.6}& \textcolor{thecolor!40!green}{98.8}& \textcolor{thecolor!40!green}{99.3}&\textcolor{thecolor!40!green}{95.3}\\
     \specialrule{.13em}{.05em}{.05em}
\end{tabular}
\end{table*}

%% file: main.bbl
\begin{thebibliography}{42}
\providecommand{\natexlab}[1]{#1}
\providecommand{\url}[1]{\texttt{#1}}
\expandafter\ifx\csname urlstyle\endcsname\relax
  \providecommand{\doi}[1]{doi: #1}\else
  \providecommand{\doi}{doi: \begingroup \urlstyle{rm}\Url}\fi

\bibitem[Andriluka et~al.(2014)Andriluka, Pishchulin, Gehler, and Schiele]{MPII}
Mykhaylo Andriluka, Leonid Pishchulin, Peter Gehler, and Bernt Schiele.
\newblock 2d human pose estimation: New benchmark and state of the art analysis.
\newblock In \emph{2014 IEEE Conference on Computer Vision and Pattern Recognition}, pages 3686--3693, 2014.

\bibitem[Arnab et~al.(2019)Arnab, Doersch, and Zisserman]{arnab2019exploiting}
Anurag Arnab, Carl Doersch, and Andrew Zisserman.
\newblock Exploiting temporal context for 3d human pose estimation in the wild.
\newblock In \emph{Proceedings of the IEEE/CVF Conference on Computer Vision and Pattern Recognition}, pages 3395--3404, 2019.

\bibitem[Black et~al.(2023)Black, Patel, Tesch, and Yang]{BEDLAM}
Michael~J. Black, Priyanka Patel, Joachim Tesch, and Jinlong Yang.
\newblock Bedlam: A synthetic dataset of bodies exhibiting detailed lifelike animated motion.
\newblock In \emph{Proceedings of the IEEE/CVF Conference on Computer Vision and Pattern Recognition (CVPR)}, pages 8726--8737, 2023.

\bibitem[Blackman(2014)]{mixamo}
Sue Blackman.
\newblock Rigging with mixamo.
\newblock \emph{Unity for Absolute Beginners}, pages 565--573, 2014.

\bibitem[Cai et~al.(2019)Cai, Ge, Liu, Cai, Cham, Yuan, and Thalmann]{Spatio-Temp}
Yujun Cai, Liuhao Ge, Jun Liu, Jianfei Cai, Tat-Jen Cham, Junsong Yuan, and Nadia~Magnenat Thalmann.
\newblock Exploiting spatial-temporal relationships for 3d pose estimation via graph convolutional networks.
\newblock In \emph{Proceedings of the IEEE/CVF international conference on computer vision}, pages 2272--2281, 2019.

\bibitem[Cao et~al.(2017)Cao, Simon, Wei, and Sheikh]{cao2017realtime}
Zhe Cao, Tomas Simon, Shih-En Wei, and Yaser Sheikh.
\newblock Realtime multi-person 2d pose estimation using part affinity fields.
\newblock In \emph{Proceedings of the IEEE conference on computer vision and pattern recognition}, pages 7291--7299, 2017.

\bibitem[Chen et~al.(2018)Chen, Wang, Peng, Zhang, Yu, and Sun]{CPN}
Yilun Chen, Zhicheng Wang, Yuxiang Peng, Zhiqiang Zhang, Gang Yu, and Jian Sun.
\newblock Cascaded pyramid network for multi-person pose estimation.
\newblock In \emph{Proceedings of the IEEE conference on computer vision and pattern recognition}, pages 7103--7112, 2018.

\bibitem[Cheng et~al.(2019)Cheng, Yang, Wang, Wending, and Tan]{OcclAware}
Yu Cheng, Bo Yang, Bo Wang, Yan Wending, and Robby Tan.
\newblock Occlusion-aware networks for 3d human pose estimation in video.
\newblock In \emph{2019 IEEE/CVF International Conference on Computer Vision (ICCV)}, pages 723--732, 2019.

\bibitem[Cheng et~al.(2021)Cheng, Wang, Yang, and Tan]{Graph&TempCNN}
Yu Cheng, Bo Wang, Bo Yang, and Robby~T Tan.
\newblock Graph and temporal convolutional networks for 3d multi-person pose estimation in monocular videos.
\newblock In \emph{Proceedings of the AAAI Conference on Artificial Intelligence}, pages 1157--1165, 2021.

\bibitem[Community(2018)]{blender_off}
Blender~Online Community.
\newblock \emph{Blender - a 3D modelling and rendering package}.
\newblock Blender Foundation, Stichting Blender Foundation, Amsterdam, 2018.

\bibitem[Fang et~al.(2017)Fang, Xie, Tai, and Lu]{fang2017rmpe}
Hao-Shu Fang, Shuqin Xie, Yu-Wing Tai, and Cewu Lu.
\newblock Rmpe: Regional multi-person pose estimation.
\newblock In \emph{Proceedings of the IEEE international conference on computer vision}, pages 2334--2343, 2017.

\bibitem[He et~al.(2017)He, Gkioxari, Doll{\'a}r, and Girshick]{he2017mask}
Kaiming He, Georgia Gkioxari, Piotr Doll{\'a}r, and Ross Girshick.
\newblock Mask r-cnn.
\newblock In \emph{Proceedings of the IEEE international conference on computer vision}, pages 2961--2969, 2017.

\bibitem[Hu et~al.(2021)Hu, Zhang, Zhan, Zhang, and Wong]{directGCN}
Wenbo Hu, Changgong Zhang, Fangneng Zhan, Lei Zhang, and Tien-Tsin Wong.
\newblock Conditional directed graph convolution for 3d human pose estimation.
\newblock In \emph{Proceedings of the 29th ACM International Conference on Multimedia}, pages 602--611, 2021.

\bibitem[Ionescu et~al.(2014)Ionescu, Papava, Olaru, and Sminchisescu]{6682899}
Catalin Ionescu, Dragos Papava, Vlad Olaru, and Cristian Sminchisescu.
\newblock Human3.6m: Large scale datasets and predictive methods for 3d human sensing in natural environments.
\newblock \emph{IEEE Transactions on Pattern Analysis and Machine Intelligence}, 36\penalty0 (7):\penalty0 1325--1339, 2014.

\bibitem[Iqbal et~al.(2017)Iqbal, Milan, and Gall]{PoseTrack}
Umar Iqbal, Anton Milan, and Juergen Gall.
\newblock Posetrack: Joint multi-person pose estimation and tracking.
\newblock In \emph{Proceedings of the IEEE Conference on Computer Vision and Pattern Recognition}, pages 2011--2020, 2017.

\bibitem[Joo et~al.(2015)Joo, Liu, Tan, Gui, Nabbe, Matthews, Kanade, Nobuhara, and Sheikh]{joo2015panoptic}
Hanbyul Joo, Hao Liu, Lei Tan, Lin Gui, Bart Nabbe, Iain Matthews, Takeo Kanade, Shohei Nobuhara, and Yaser Sheikh.
\newblock Panoptic studio: A massively multiview system for social motion capture.
\newblock In \emph{Proceedings of the IEEE International Conference on Computer Vision}, pages 3334--3342, 2015.

\bibitem[Kipf and Welling(2016)]{GCN}
Thomas~N Kipf and Max Welling.
\newblock Semi-supervised classification with graph convolutional networks.
\newblock \emph{arXiv preprint arXiv:1609.02907}, 2016.

\bibitem[Lea et~al.(2017)Lea, Flynn, Vidal, Reiter, and Hager]{TCN}
Colin Lea, Michael~D. Flynn, Rene Vidal, Austin Reiter, and Gregory~D. Hager.
\newblock Temporal convolutional networks for action segmentation and detection.
\newblock In \emph{Proceedings of the IEEE Conference on Computer Vision and Pattern Recognition (CVPR)}, 2017.

\bibitem[Lee and Chen(1985)]{lee1985determination}
Hsi-Jian Lee and Zen Chen.
\newblock Determination of 3d human body postures from a single view.
\newblock \emph{Computer Vision, Graphics, and Image Processing}, 30\penalty0 (2):\penalty0 148--168, 1985.

\bibitem[Li et~al.(2018)Li, Zhou, Li, and Liu]{li2018bottom}
Miaopeng Li, Zimeng Zhou, Jie Li, and Xinguo Liu.
\newblock Bottom-up pose estimation of multiple person with bounding box constraint.
\newblock In \emph{2018 24th international conference on pattern recognition (ICPR)}, pages 115--120. IEEE, 2018.

\bibitem[Li and Chan(2015)]{li20153d}
Sijin Li and Antoni~B Chan.
\newblock 3d human pose estimation from monocular images with deep convolutional neural network.
\newblock In \emph{Computer Vision--ACCV 2014: 12th Asian Conference on Computer Vision, Singapore, Singapore, November 1-5, 2014, Revised Selected Papers, Part II 12}, pages 332--347. Springer, 2015.

\bibitem[Lin et~al.(2014)Lin, Maire, Belongie, Hays, Perona, Ramanan, Doll{\'a}r, and Zitnick]{COCO}
Tsung-Yi Lin, Michael Maire, Serge Belongie, James Hays, Pietro Perona, Deva Ramanan, Piotr Doll{\'a}r, and C~Lawrence Zitnick.
\newblock Microsoft coco: Common objects in context.
\newblock In \emph{Computer Vision--ECCV 2014: 13th European Conference, Zurich, Switzerland, September 6-12, 2014, Proceedings, Part V 13}, pages 740--755. Springer, 2014.

\bibitem[Lin et~al.(2017)Lin, Doll{\'a}r, Girshick, He, Hariharan, and Belongie]{FPN}
Tsung-Yi Lin, Piotr Doll{\'a}r, Ross Girshick, Kaiming He, Bharath Hariharan, and Serge Belongie.
\newblock Feature pyramid networks for object detection.
\newblock In \emph{Proceedings of the IEEE conference on computer vision and pattern recognition}, pages 2117--2125, 2017.

\bibitem[Mahmood et~al.(2019)Mahmood, Ghorbani, Troje, Pons-Moll, and Black]{AMASS}
Naureen Mahmood, Nima Ghorbani, Nikolaus~F Troje, Gerard Pons-Moll, and Michael~J Black.
\newblock Amass: Archive of motion capture as surface shapes.
\newblock In \emph{Proceedings of the IEEE/CVF international conference on computer vision}, pages 5442--5451, 2019.

\bibitem[Martinez et~al.(2017)Martinez, Hossain, Romero, and Little]{martinez2017simple}
Julieta Martinez, Rayat Hossain, Javier Romero, and James~J Little.
\newblock A simple yet effective baseline for 3d human pose estimation.
\newblock In \emph{Proceedings of the IEEE international conference on computer vision}, pages 2640--2649, 2017.

\bibitem[O'Shea and Nash(2015)]{CNN}
Keiron O'Shea and Ryan Nash.
\newblock An introduction to convolutional neural networks.
\newblock \emph{arXiv preprint arXiv:1511.08458}, 2015.

\bibitem[Paszke et~al.(2019)Paszke, Gross, Massa, Lerer, Bradbury, Chanan, Killeen, Lin, Gimelshein, Antiga, et~al.]{pytorch}
Adam Paszke, Sam Gross, Francisco Massa, Adam Lerer, James Bradbury, Gregory Chanan, Trevor Killeen, Zeming Lin, Natalia Gimelshein, Luca Antiga, et~al.
\newblock Pytorch: An imperative style, high-performance deep learning library.
\newblock \emph{Advances in neural information processing systems}, 32, 2019.

\bibitem[Patel et~al.(2021)Patel, Huang, Tesch, Hoffmann, Tripathi, and Black]{agora}
Priyanka Patel, Chun-Hao~P. Huang, Joachim Tesch, David~T. Hoffmann, Shashank Tripathi, and Michael~J. Black.
\newblock {AGORA}: Avatars in geography optimized for regression analysis.
\newblock In \emph{Proceedings IEEE/CVF Conf.~on Computer Vision and Pattern Recognition ({CVPR})}, 2021.

\bibitem[Pavlakos et~al.(2018)Pavlakos, Zhou, and Daniilidis]{pavlakos2018ordinal}
Georgios Pavlakos, Xiaowei Zhou, and Kostas Daniilidis.
\newblock Ordinal depth supervision for 3d human pose estimation.
\newblock In \emph{Proceedings of the IEEE conference on computer vision and pattern recognition}, pages 7307--7316, 2018.

\bibitem[Pavllo et~al.(2019)Pavllo, Feichtenhofer, Grangier, and Auli]{VideoPose3D}
Dario Pavllo, Christoph Feichtenhofer, David Grangier, and Michael Auli.
\newblock 3d human pose estimation in video with temporal convolutions and semi-supervised training.
\newblock In \emph{Proceedings of the IEEE/CVF conference on computer vision and pattern recognition}, pages 7753--7762, 2019.

\bibitem[Qiu et~al.(2020)Qiu, Qiu, Fu, and Fu]{DGCN}
Zhongwei Qiu, Kai Qiu, Jianlong Fu, and Dongmei Fu.
\newblock Dgcn: Dynamic graph convolutional network for efficient multi-person pose estimation.
\newblock In \emph{Proceedings of the AAAI Conference on Artificial Intelligence}, pages 11924--11931, 2020.

\bibitem[Reddi et~al.(2019)Reddi, Kale, and Kumar]{amsgrad}
Sashank~J Reddi, Satyen Kale, and Sanjiv Kumar.
\newblock On the convergence of adam and beyond.
\newblock \emph{arXiv preprint arXiv:1904.09237}, 2019.

\bibitem[S{\'a}r{\'a}ndi et~al.(2018)S{\'a}r{\'a}ndi, Linder, Arras, and Leibe]{H36MOccl}
Istv{\'a}n S{\'a}r{\'a}ndi, Timm Linder, Kai~O Arras, and Bastian Leibe.
\newblock How robust is 3d human pose estimation to occlusion?
\newblock \emph{arXiv preprint arXiv:1808.09316}, 2018.

\bibitem[Shan et~al.(2023)Shan, Liu, Zhang, Wang, Han, Wang, Ma, and Gao]{D3DP}
Wenkang Shan, Zhenhua Liu, Xinfeng Zhang, Zhao Wang, Kai Han, Shanshe Wang, Siwei Ma, and Wen Gao.
\newblock Diffusion-based 3d human pose estimation with multi-hypothesis aggregation.
\newblock In \emph{Proceedings of the IEEE/CVF International Conference on Computer Vision (ICCV)}, pages 14761--14771, 2023.

\bibitem[Tome et~al.(2017)Tome, Russell, and Agapito]{tome2017lifting}
Denis Tome, Chris Russell, and Lourdes Agapito.
\newblock Lifting from the deep: Convolutional 3d pose estimation from a single image.
\newblock In \emph{Proceedings of the IEEE conference on computer vision and pattern recognition}, pages 2500--2509, 2017.

\bibitem[Toshev and Szegedy(2014)]{toshev2014deeppose}
Alexander Toshev and Christian Szegedy.
\newblock Deeppose: Human pose estimation via deep neural networks.
\newblock In \emph{Proceedings of the IEEE conference on computer vision and pattern recognition}, pages 1653--1660, 2014.

\bibitem[Varol et~al.(2017)Varol, Romero, Martin, Mahmood, Black, Laptev, and Schmid]{SURREAL}
Gul Varol, Javier Romero, Xavier Martin, Naureen Mahmood, Michael~J. Black, Ivan Laptev, and Cordelia Schmid.
\newblock Learning from synthetic humans.
\newblock In \emph{2017 IEEE Conference on Computer Vision and Pattern Recognition (CVPR)}. IEEE, 2017.

\bibitem[Wojke et~al.(2017)Wojke, Bewley, and Paulus]{DeepSort}
Nicolai Wojke, Alex Bewley, and Dietrich Paulus.
\newblock Simple online and realtime tracking with a deep association metric.
\newblock In \emph{2017 IEEE international conference on image processing (ICIP)}, pages 3645--3649. IEEE, 2017.

\bibitem[Wu et~al.(2019)Wu, Kirillov, Massa, Lo, and Girshick]{wu_kirillov_massa_lo_girshick_2019}
Yuxin Wu, Alexander Kirillov, Francisco Massa, Wan-Yen Lo, and Ross Girshick.
\newblock Detectron2: A pytorch-based modular object detection library.
\newblock \emph{Meta AI}, 10, 2019.

\bibitem[Zhao et~al.(2023)Zhao, Zheng, Liu, Wang, and Chen]{PFV2}
Qitao Zhao, Ce Zheng, Mengyuan Liu, Pichao Wang, and Chen Chen.
\newblock Poseformerv2: Exploring frequency domain for efficient and robust 3d human pose estimation.
\newblock In \emph{Proceedings of the IEEE/CVF Conference on Computer Vision and Pattern Recognition}, pages 8877--8886, 2023.

\bibitem[Zheng et~al.(2021)Zheng, Zhu, Mendieta, Yang, Chen, and Ding]{PF}
Ce Zheng, Sijie Zhu, Matias Mendieta, Taojiannan Yang, Chen Chen, and Zhengming Ding.
\newblock 3d human pose estimation with spatial and temporal transformers.
\newblock In \emph{Proceedings of the IEEE/CVF International Conference on Computer Vision}, pages 11656--11665, 2021.

\bibitem[Zheng et~al.(2023)Zheng, Wu, Chen, Yang, Zhu, Shen, Kehtarnavaz, and Shah]{survey}
Ce Zheng, Wenhan Wu, Chen Chen, Taojiannan Yang, Sijie Zhu, Ju Shen, Nasser Kehtarnavaz, and Mubarak Shah.
\newblock Deep learning-based human pose estimation: A survey.
\newblock \emph{ACM Computing Surveys}, 56\penalty0 (1):\penalty0 1--37, 2023.

\end{thebibliography}
